# Critique of Impure Reason: Unveiling the reasoning behaviour of medical Large Language Models

**Shamus Sim Zi Yang**
QueueMed Healthtech
Kuala Lumpur
shamus@qmed.asia
Both authors contributed equally to this work
Correspondence can be addressed to both authors

**Tyrone Chen**
Peter MacCallum Cancer Centre
Melbourne*
tyrone.chen@petermac.org.au
Both authors contributed equally to this work
Correspondence can be addressed to both authors
*This project was conducted independent of this author's institution and details are provided for correspondence only*

## Abstract

**Background:** Despite the current ubiquity of Large Language Models (LLMs) across the medical domain, there is a surprising lack of studies which address their *reasoning behaviour*. We emphasise the importance of understanding *reasoning behaviour* as opposed to high-level prediction accuracies, since it is equivalent to explainable AI (XAI) in this context. In particular, achieving XAI in medical LLMs used in the clinical domain will have a significant impact across the healthcare sector. **Results:** Therefore, in this work, we adapt the existing concept of *reasoning behaviour* and articulate its interpretation within the specific context of medical LLMs. We survey and categorise current state-of-the-art approaches for modeling and evaluating *reasoning reasoning* in medical LLMs. Additionally, we propose theoretical frameworks which can empower medical professionals or machine learning engineers to gain insight into the low-level reasoning operations of these previously obscure models. We also outline key open challenges facing the development of *Large Reasoning Models*. **Conclusion:** The subsequent increased transparency and trust in medical machine learning models by clinicians as well as patients will accelerate the integration, application as well as further development of medical AI for the healthcare system as a whole.

# 1 Introduction

**Key points:**

- Many medical LLM studies achieve high-level benchmarks but do not explore fundamental *reasoning behaviour*.
- Exposing *reasoning behaviour* in medical LLMs increases transparency and subsequently trust by users and beneficiaries.
- We evaluate the current state-of-the-art of exploring *reasoning behaviour* in medical LLM, and propose several actionable frameworks.
- We discuss current open challenges in building *Large Reasoning Models* for the medical field.

*Reasoning* drives problem solving activities, and is ubiquitous in our daily lives. The rising adoption of the field of artificial intelligence and its proximity to the concept of reasoning then naturally provokes the question: what is the *reasoning behaviour* of machine learning models commonly used in artificial intelligence [Figure 1]? This is particularly pertinent with regard to the increasing use of large language models (LLMs).

In this review, we focus specifically on **transformer-based large language models (LLMs)**, which are built on the transformer architecture, an attention-based mechanism capable of capturing complex dependencies in sequential data [1]. These models are characterized by a large number of trainable parameters and are pre-trained on vast corpora of textual data.

We define **medical LLMs** in an application-centric manner, referring broadly to any LLMs that are employed as a core component in medical or clinical workflows. This broad definition accommodates a diverse range of use cases, including applications in diagnosis [2, 3, 4], medical image analysis [5, 6], clinical summarisation [7] and EHR question-answering tasks [8]. This includes both domain-specific models pre-trained or fine-tuned on biomedical corpora, as well as general-purpose LLMs (e.g., GPT or LLAMA models) that are adapted or prompted for use in medical settings. While such general models may not have been originally developed for healthcare, they are increasingly leveraged in a wide range of medical applications and thus fall within the scope of our discussion. In this review, we observed the frequent use of various variants of GPT- and LLaMA-based models throughout the literature. Due to their sheer scale and complexity, LLMs are inherently less interpretable than traditional machine learning models, which are themselves often considered "black boxes" under normal circumstances. We note, however, that although LLMs have also been applied directly to genomic and molecular sequence data [9] with potential medical use, the present review does not cover these applications.

The question of *how* LLMs arrive at their answers—particularly in high-stakes applications like medicine—is surprisingly underexplored. This gap is especially striking given the widespread deployment of these models across domains, often without a comprehensive understanding of their underlying reasoning mechanisms. Instead, evaluations tend to focus on performance metrics such as accuracy, F1 scores, precision, and recall. These metrics are typically benchmarked against curated subsets of state-of-the-art models as well as specialised datasets which consist of medical license examination questions from the United States (USMLE), Mainland China (MCMLE), Taiwan (TWMLE) [10], India (AIIMS/NEET) [11] and other broader questions less directly related to clinical fields [12, 13]. While these may be effective in some cases, such metrics provide limited insight into the complex and obscure inferential processes that LLMs apply to generate answers.

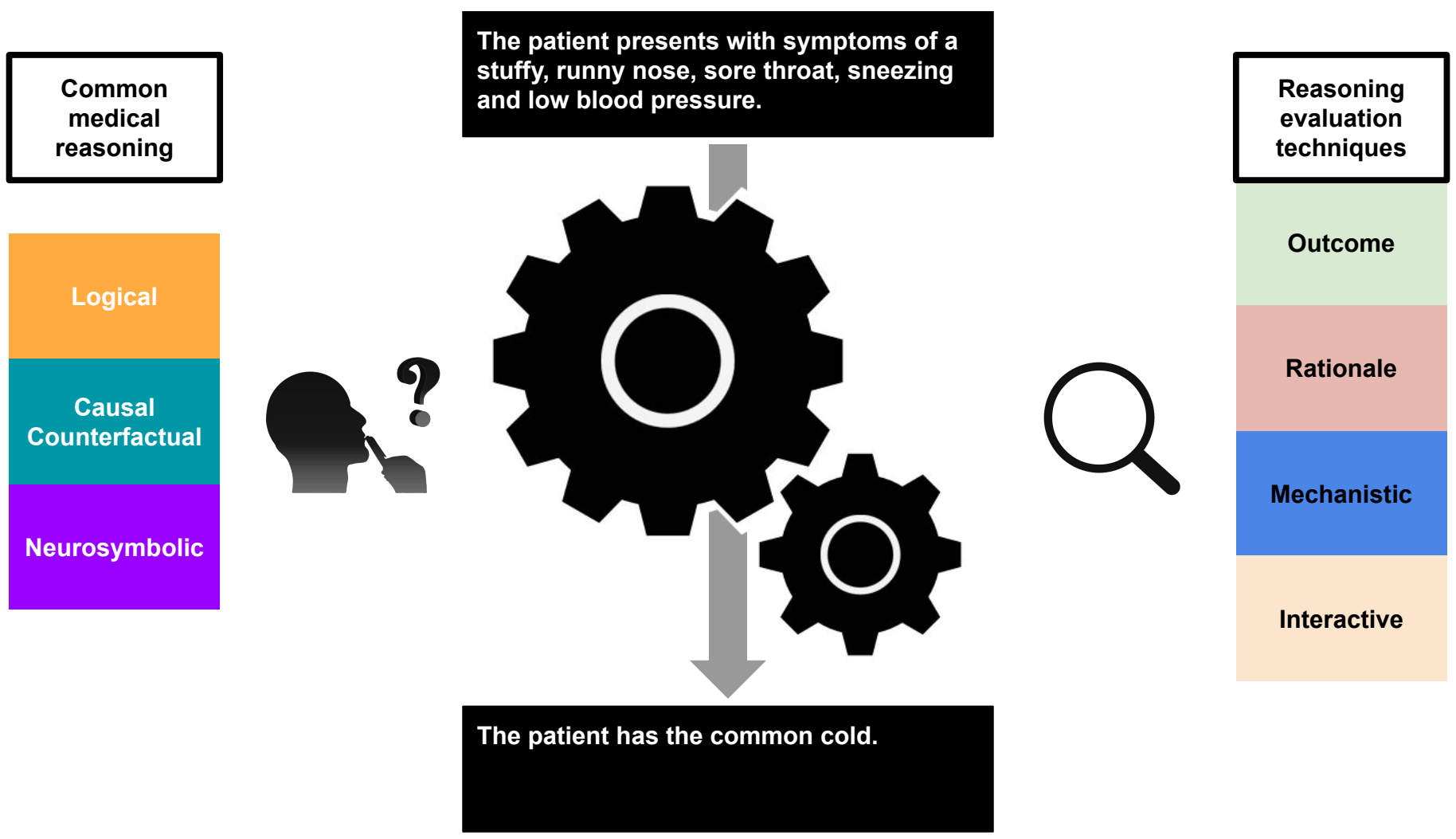

Figure 1: A graphical abstract illustrating the current state of medical LLMs in the context of *reasoning behaviour*.

This neglect in understanding *reasoning behaviour* leads to their unintentional misuse with direct real-world effects, including data fabrication [14], false accusations [15] and suicide [16]. Further obfuscating *reasoning behaviour* in LLMs, particularly generative models, is their ability to mimic the semantics of question and answer processes convincingly while being surprisingly accurate, to the point where individuals have mistakenly assumed their sentience [17]. Such issues are amplified in LLMs used for medical purposes, given their proximity to life-and-death decisions, for example in the case of acute heart failure [18].

We highlight the importance of gaining insight into the process-driven, *reasoning behaviour* by observing its functional similarity to *interpretability metrics* in machine learning [19, 20, 21, 22, 23]. In both cases the aim is to inspect the learning process of the model, with corresponding metrics used to gain information into how a model is correctly or incorrectly predisposed towards a certain outcome, in an attempt to address the common "black-box" problem of machine learning models [Figure 2]. We re-emphasise the lack of LLM studies tackling the question of general inference to focus on a noticeable gap in the field: **remarkably few studies investigate *reasoning behaviour* in the medical LLM space.**

Given the increased stakes of medical LLMs in clinical decision making, achieving a deeper understanding of medical LLMs carries a greater weight than with general-purpose LLMs. Thus, their intense scrutiny by both medical experts and the general public is unsurprising. Therefore, it is necessary to supplement clinicians with insight into the *reasoning behaviour* of medical LLMs to better understand how they arrive at their conclusions and expose potential logical fallacies. An ability for LLMs to provide reasoning for their outputs, for example, in a medical recommendation or diagnosis, allows clinicians to clarify discrepancies between machine and expert suggestions. This transparency fosters trust, encouraging integration of LLMs and other machine learning models into clinical decisions and subsequently improving patient outcomes.

In our review, we will address a few specific points:

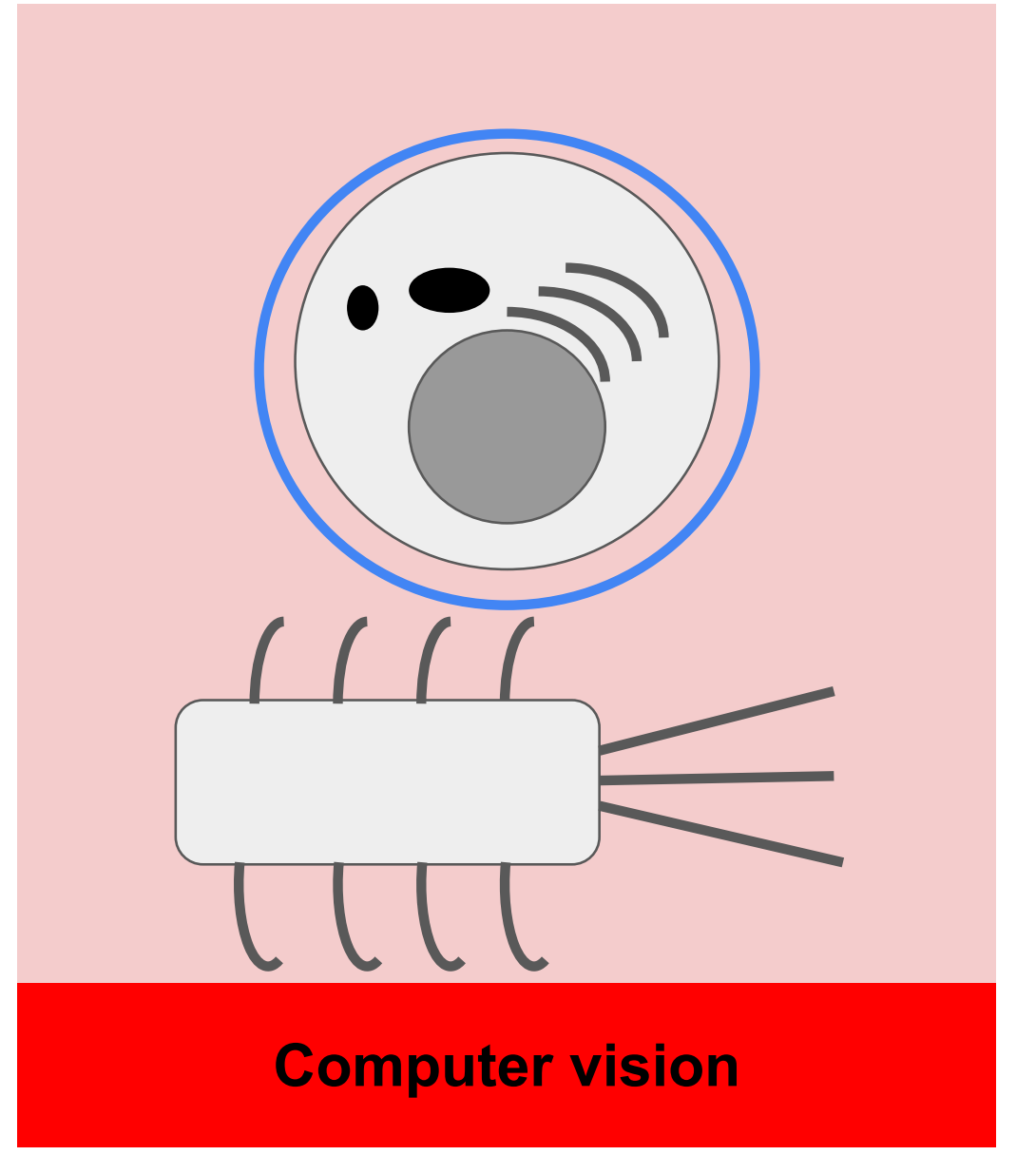

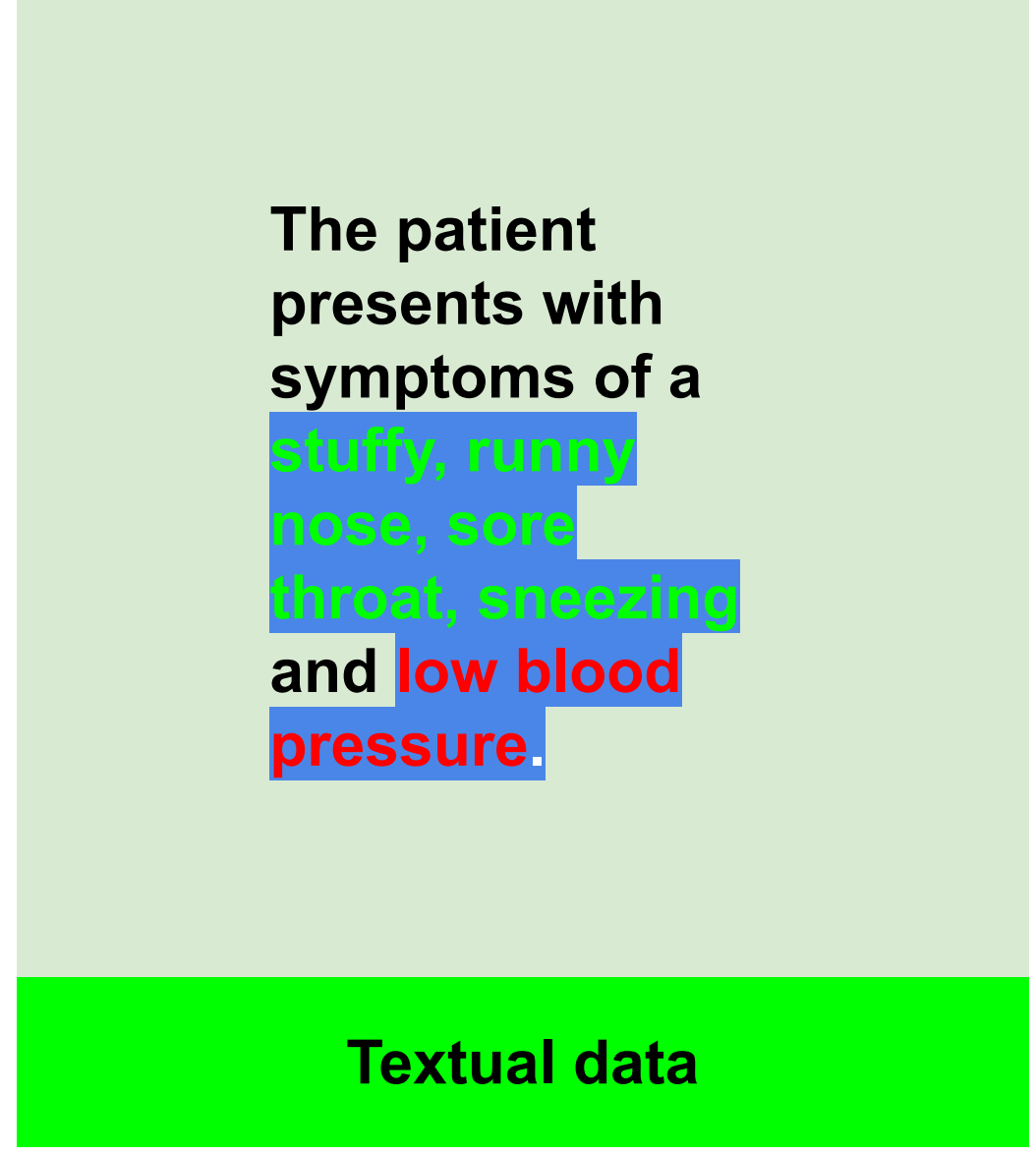

Figure 2: An illustration of the contrast in modalities between computer vision and natural language processing.

1. We provide a primer introducing fundamental AI concepts covered in this review.
2. We adopt the existing concept of *reasoning behaviour* and articulate its interpretation within the specific context of medical LLMs.
3. We discuss the importance of evaluating *reasoning behaviour* in addition to performance metrics
4. We compare and contrast the current state-of-the-art (SOTA) in *reasoning behaviour* for the medical field, and note a surprising lack of such studies
5. We propose strategies to improve and evaluate the *reasoning behaviour* of medical LLMs, which will grant greater transparency

# 2 Primer of Foundational Concepts in Reasoning

## 2.1 Prompting methods for Reasoning

Prompting methods are lightweight techniques that guide large language models (LLMs) to perform more structured reasoning without additional training. In this section, we focus on two representative prompting-based strategies: Chain-of-Thought (CoT) prompting, which encourages sequential reasoning through intermediate steps, and Tree-of-Thought (ToT) prompting, which allows the model to explore multiple reasoning paths through structured search and self-evaluation.

### 2.1.1 Chain of Thought Reasoning

Chain-of-Thought (CoT) [Figure 3] reasoning is a prompting technique used to improve an LLM's ability to solve reasoning-related problems. Rather than producing an answer directly, the model decomposes the problem into a sequence of intermediate steps, thereby facilitating more structured and transparent reasoning [24]. However, in some cases, LLMs may still generate incorrect intermediate steps [25].

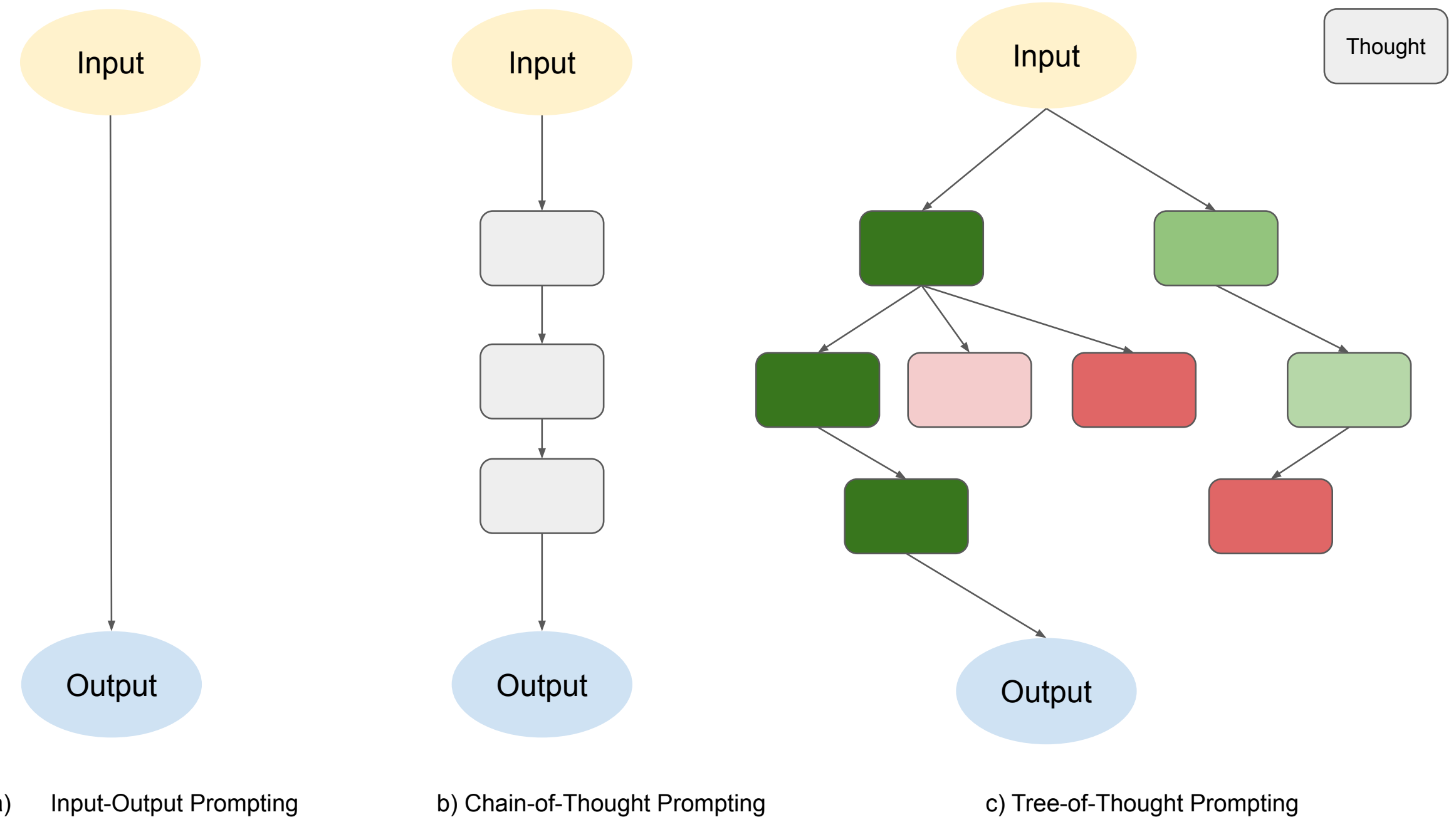

Figure 3: A schematic diagram illustrating different strategies for solving problems using LLMs. Each rectangular box represents a "thought"—a meaningful segment of language that functions as an intermediate step in the reasoning or problem-solving process.

#### 2.1.2 Tree of Thought Reasoning

Tree of Thoughts (ToT) [Figure 3] is an advanced reasoning framework that extends the problem solving capabilities of large language models (LLMs). [26]. Instead of token-by-token generating responses in a linear fashion, ToT prompts the model to generate coherent reasoning steps or "thoughts," which are organized into a tree structure [26]. Each node in this tree represents a partial solution, and the model can explore multiple possible continuations (branching), evaluate their promise using heuristic reasoning (state evaluation), and make decisions through structured search strategies such as breadth-first or depth-first search [26].This modular approach includes four core components: decomposing problems into thought steps, generating multiple candidate thoughts, heuristically evaluating them, and using search algorithms to navigate the tree [26]. By combining these elements, ToT enables LLMs to reason with greater depth, backtrack to previous nodes when necessary, and make more globally informed decisions, features that are particularly relevant for complex biomedical tasks such as differential diagnosis or clinical planning.

### 2.2 Agentic-based Methods

**Agent-based methods** for LLM reasoning are an emerging paradigm designed to enhance the problem-solving capabilities of large language models (LLMs) by structuring their operations as autonomous or semi-autonomous agents. These approaches address the limitations of standard LLM prompting by introducing explicit planning, memory, iterative decision-making, and tool usage.

In the medical domain, agent-based methods leverage LLMs as modular agents, each assigned a specific function. Examples include clinical triage, medical literature retrieval, summarization of patient data [7], decision support, or guideline compliance checking.

Common features of agent-based medical reasoning with LLMs include:

- **Iterative Planning:** The model decomposes complex clinical problems into sub-questions, enabling step-by-step analysis and dynamic adjustments as new information becomes available [27].
- **Memory Integration:** Some agents maintain both short- and long-term memory, allowing them to track patient history, previous actions, and evolving diagnostic hypotheses over time [27].
- **Tool Augmentation:** LLM agents can interact with external tools—such as medical databases like EHRs [8], drug databases [28] or medical knowledge graphs [28] medical calculators [29], or literature search engines—to retrieve up-to-date information and perform specialized computations.
- **Reflection**: Agents incorporate feedback mechanisms [30] and reflective decision-making [28] to revise and improve reasoning dynamically, reducing hallucinations and adapting to new inputs.
- **Multi-agent collaborative Group Reasoning**: A multiagent reasoning framework involves deploying multiple specialized LLM-based agents [30, 28]—such as efficacy, safety, or diagnostic agents [28]—that collaboratively analyze clinical information, challenge each other's conclusions, and synthesize decisions through structured dialogue or argumentation [30]. This mirrors multidisciplinary clinical teams and enhances reasoning transparency, robustness, and safety.

### 2.3 Learning based approaches for reasoning: Supervised to Reinforcement Learning

LLMs can be prompted into producing CoT- sequence of tokens representing intermediate steps in the reasoning process. However, LLMs lack explicit training objectives that encourage deep, stepwise CoT before arriving at a final answer. Large Reasoning Models (LRMs), a subclass of LLMs close this gap by being trained to perform extended reasoning in CoT, before taking actions or producing final a final answer. The release of OpenAI's o1 series exemplifies this emerging trajectory.

In the emerging landscape of *learning-to-reason* in LRMs, the training process often begins with **Supervised Fine-Tuning (SFT)**, where models are trained on labeled reasoning datasets to capture task-specific logic and patterns. However, as reasoning tasks grow more nuanced and open-ended, SFT alone becomes insufficient. To overcome these limitations, **Reinforcement Learning from Human Feedback (RLHF)** has been introduced. RLHF refines model output by training a reward model based on human preferences, enabling the LLM to generate responses that are more aligned with human-like reasoning: coherent, responsible and contextually appropriate.

Beyond training, recent research also demonstrates the benefits of *inference-time scaling*, sometimes also know as *test-time scaling* [31, 32], where prompting LLMs to generate longer or multiple reasoning paths (e.g., via tree-of-thought prompting) can significantly improve inference-time performance. These advances in both training and inference signal a shift toward what has been described as *Large Reasoning Models*—LLMs specifically optimized for robust, interpretable, and multi-step reasoning in CoT.

In the sections that follow, we describe two foundational learning strategies—Supervised Fine-Tuning (SFT) and Reinforcement Learning from Human Feedback (RLHF)—both of which have been shown to align LLM outputs more closely with expert medical reasoning and improve clinical utility.

#### 2.3.1 Supervised Fine-tuning

Supervised fine-tuning (SFT) is a process used to improve a pre-trained AI model, such as a large language model, so it performs better on specific tasks or domains [33]. It involves training the model on a carefully prepared dataset that includes input examples paired with the correct outputs (labels). This helps the model learn the desired behavior more precisely by adjusting its parameters based on these examples.

The key steps in supervised fine-tuning are [33]:

1. Starting with a pre-trained model that already understands general language or knowledge.
2. Preparing a labeled dataset relevant to the specific task, where each input has a correct output.
3. Training the model on this dataset to fine-tune its parameters, so it responds accurately in the targeted context.
4. Evaluating and iterating to ensure the model improves without overfitting.

This method transforms a general-purpose model into a specialized model that delivers better performance on domain-specific tasks [33].

#### 2.3.2 Reinforcement Learning with Human Feedback

Reinforcement Learning from Human Feedback (RLHF) [34]in the context of LLMs, is a technique used to teach LLMs to behave in ways that align better with human preferences [33, 35]. Instead of relying solely on fixed rules or labeled data, RLHF uses human feedback to guide the learning process.

The standard RLHF pipeline has three stages: [36] [35] ) :

1. **Supervised Fine-Tuning (Optional):** Often one starts with an LLM that's been fine-tuned on high-quality responses with reasoning traces to a some sample of prompts(datasets).
2. **Reward Model Training:**
   - Humans see multiple candidate outputs for the same prompt.
   - They rank or compare outputs by which one they prefer.
   - A small neural network (*reward model*) is trained to predict those human preference scores as a single scalar reward.
3. **Policy Optimization:**
   - The LLM (now called the *policy*) is fine-tuned with a reinforcement-learning algorithm (e.g. PPO).
   - At each update, the policy generates outputs, the reward model scores them, and the RL algorithm nudges the policy toward higher-reward actions.

Over successive iterations, the model learns to produce answers that are more helpful, more accurate, and more aligned with human values [36, 35]. As an example of its applicability, RLHF underpins the effectiveness of systems like GPT-4 [35].

There are several variants of the RL step (PPO [37], DPO [38], GRPO [39], etc.), each with its own trade-offs in stability, efficiency, and memory usage. We won't dive into their details here; instead, our focus will be on how this human-in-the-loop process is key to building LLMs that can carry out complex medical reasoning tasks.

### 2.4 Directed Acyclic Graphs

A Directed Acyclic Graph (DAG) is a graphical structure made up of nodes (also called vertices) connected by arrows (edges) that indicate a specific direction of influence from one variable to another [40]. The term "acyclic" means that the graph contains no closed loops—following the arrows from node to node will never bring you back to the starting point [40]. This enforces a clear, unidirectional flow of information, which is essential for modeling causal relationships.

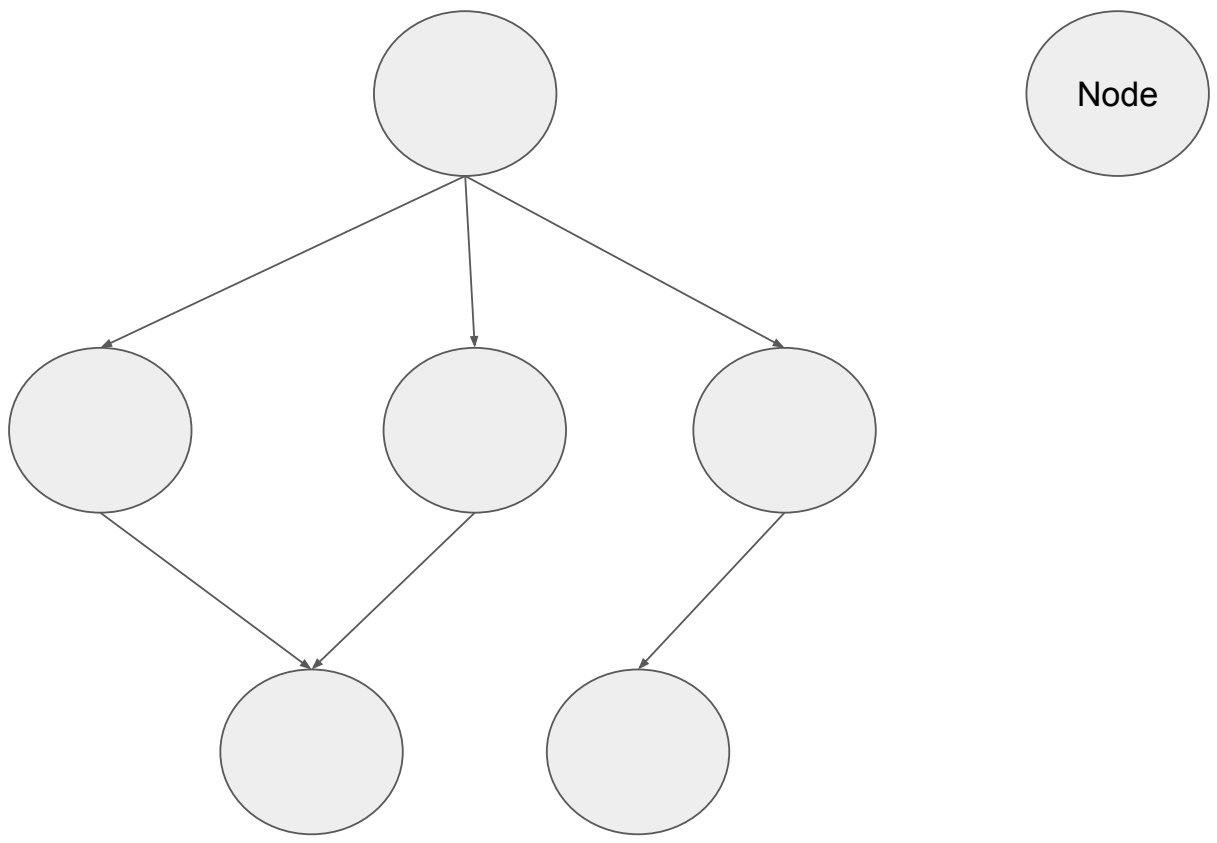

Figure 4: A sample Directed Acyclic Graph.

DAGs are commonly used to represent processes or systems where order matters and repetition is not allowed [40]. For example, DAGs can clarify whether the observed link between paracetamol use and childhood wheezing is due to a true causal effect or confounded by factors like viral infections [41].

This structure is particularly useful in biomedical AI for modeling causal relationships [41], clinical reasoning pathways [42], or decision-making logic in a transparent and interpretable way [40, 43, 44]. ref to other dag papers here

# 3 What is *reasoning behaviour* in the Context of Medical LLMs?

First, we specifically define *reasoning behaviour* in the context of our review. It is important to note that the general term *reasoning* is used loosely across LLM-related literature, and often *reasoning behaviour* is not the focus of the experiment but high-level performance metrics are. Here, we slightly adapt the specific definitions of *reasoning* and *reasoning behaviour* respectively from Mondorf & Plank [45], who define these concepts in the context of general LLMs. We also add a third definition: *reasoning outcome*.

*Reasoning: "The process of drawing conclusions based on available information."*

*Reasoning outcome: "An event where reasoning reaches a conclusion."*

*Reasoning behaviour: "The specific flow of logic within the scope of available information in a system that leads to a reasoning outcome."*

Specifically, while the *outcome* of *reasoning* is an event where a conclusion is obtained, *reasoning behaviour* describes the *process of how* the conclusion is obtained [Figure 5]. The vast majority of generic and medical-LLMs focus on the former while disregarding the latter.

We apply the same definition to this review for medical-LLMs.

# 4 Types of Reasoning Applicable to Medical LLMs

This section reviews studies that extend beyond a high-level focus on task accuracy, focusing instead on evaluating the *reasoning behaviour* of large language models. *Reasoning behaviour* can be subdivided into multiple categories [Table 4]. However, for the purposes of this study, we focus mainly on subtypes of logical reasoning [46] and causal reasoning [47] that are common in medical-LLMs. In addition, we explore the less visible field of neurosymbolic reasoning. We note that other reasoning types such as mathematical reasoning [48] may be more applicable to other categories of LLMs, which are not the focus of this review.

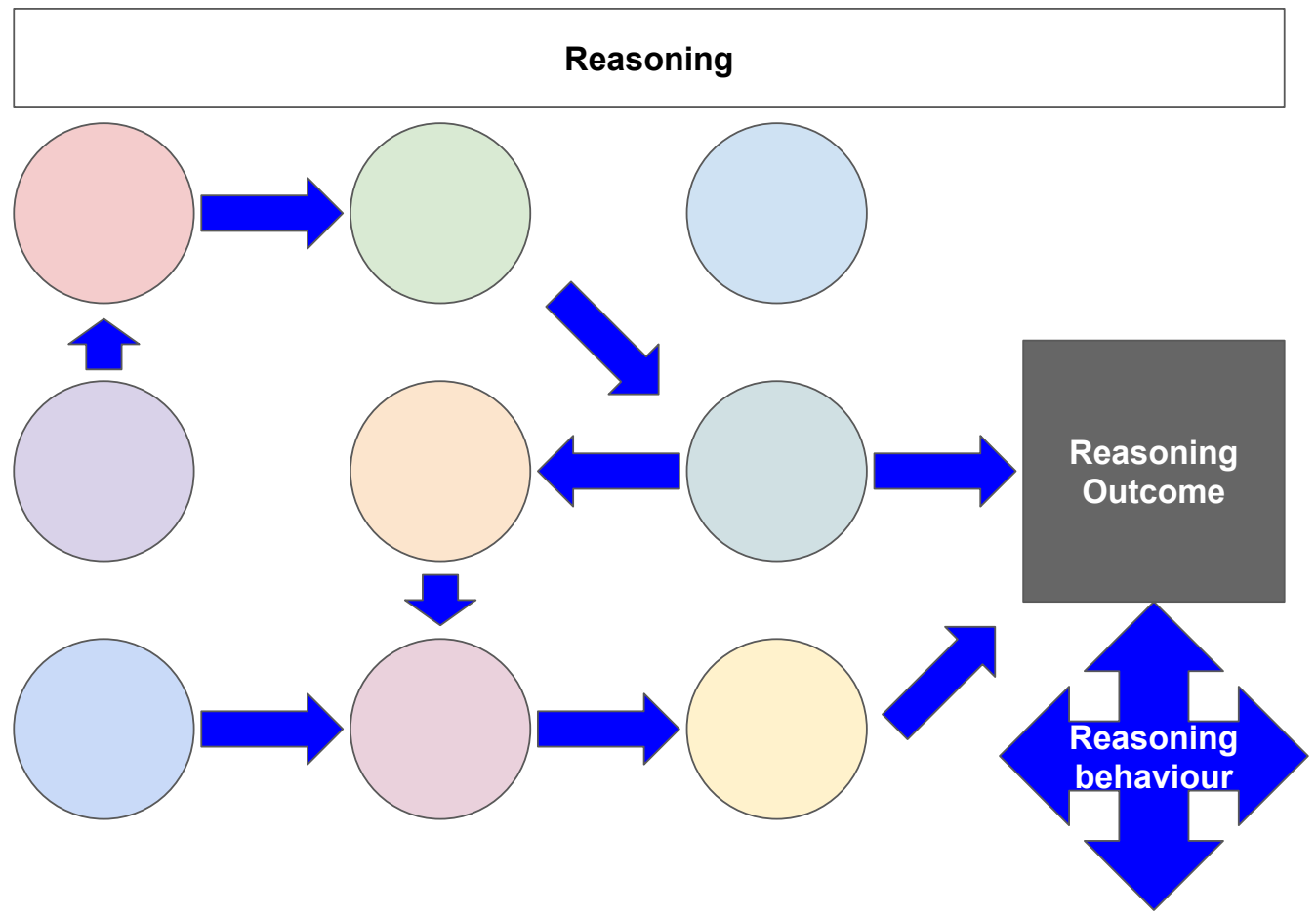

Figure 5: A graphical representation of *reasoning*, *reasoning outcome* and *reasoning behaviour*. *Reasoning* encapsulates the process of drawing conclusions, arriving at a *reasoning outcome*. At a more fundamental level, *reasoning behaviour* describes the logical flow through the system that occurs during *reasoning*

### 4.1 Logical Reasoning

The study of logical reasoning addresses the question of how individuals infer valid conclusions from a set of given premises within a structured framework of logical rules and principles [45]. Mondorf & Plank [45] classify logical reasoning into: deductive, inductive, and abductive reasoning. Deductive and inductive reasoning both work towards a general conclusion, with the key distinction being that deductive reasoning begins with a premise while inductive reasoning begins with observations. On a broader scale, abductive reasoning involves formulating plausible hypotheses that explain incomplete observations. The key distinction between deductive reasoning and both inductive/abductive reasoning is that deductive reasoning results in clear conclusions, while inductive/abductive reasoning may not necessarily achieve this. For more nuanced and low-level details on the distinction between the three types of logical reasoning in the context of LLMs, we refer the reader to other publications [45, 49, 50].

### 4.2 Causal/Counterfactual Reasoning

Causal reasoning refers to the ability to connect cause and effect in scenarios. In the context of medical and general LLMs, their capabilities are a matter of debate. Intuitively, providing cause and effect relationships improve one's understanding of a system. Correspondingly, providing this information to medical LLMs would improve a model's "understanding" and has unsurprisingly emerged as an area of interest. This capability is essential in applications like medical diagnosis, where identifying causal links—such as between symptoms and potential conditions—can inform accurate and actionable insights. Causal reasoning involves not just recognising associations but distinguishing *directional* influences. In theory, knowing directionality grants the model the ability to infer, for instance, whether "A causes B" or "B causes A". In real-world medical applications, larger LLMs like GPT-4.0 are capable of inferring causal direction between variables, allowing accurate diagnosis of neuropathic pain [51].

### 4.3 Symbolic Reasoning

**Symbolic reasoning**—also known as *symbolic AI* or "Good Old-Fashioned Artificial Intelligence" — is a process that involves the use of mathematical symbols to represent concepts, objects, or relationships in order to facilitate reasoning, problem-solving, and decision-making. Unlike Machine Learning which relies on learning from vast datasets, this form of reasoning is characterised by its reliance on formal logic and structured representations, allowing for the manipulation of abstract symbols according to defined rules without requiring vast datasets. Symbolic systems execute explicit inference chains—for example, "if symptom A and test result B, then diagnosis C"—mirroring the logic of clinical guidelines and expert systems such as MYCIN [52] or INTERNIST [53]. This approach enhances interpretability and trustworthiness in biomedical applications, since the reasoning steps are transparent and auditable.

Table 1: A table showing types of reasoning, their definition and examples. Reasoning types are colour-coded for clarity. Logical reasoning encompasses abductive, deductive, and inductive subtypes. *We note that the term *symbolic reasoning* may be misleading, as it is fundamentally an abstract *data representation* which simplifies the process of translating a scenario into a *reasoning* framework.

| Type of Reasoning | Definition | General Example | Medical Example |
|---|---|---|---|
| Abductive | Inferring the most likely explanation for observed data or evidence. | Ali, Muthu and Ah Hock breathe oxygen. Therefore, Ali, Muthu and Ah Hock are likely human. | A patient has increased intercranial pressure, blurred vision and nausea. Therefore, the patient may have a brain aneurysm or ischemic stroke. |
| Deductive | Reasoning from a set of premises to reach a certain conclusion. | All humans breathe oxygen. Rentap is human. Therefore, Rentap breathes oxygen. | A patient has increased intercranial pressure, blurred vision and nausea. A CT scan shows no bleeding or swelling. Therefore, the patient does not have a brain aneurysm. |
| Inductive | Inferring general principles based on specific observations. | All humans that I have seen breathe oxygen. Therefore, Rentap probably breathes oxygen. | A patient has increased intercranial pressure, blurred vision and nausea. A CT scan shows no bleeding or swelling. Therefore, the patient probably has an ischemic stroke. |
| Symbolic* | The abstraction of a system into its component parts, which enables a more direct application of mathematics. | Rule: If an organism breathes oxygen and nitrogen, and exhales carbon dioxide → likely human.<br>Observation: Ali, Muthu, Ah Hock, and Rentap exhibit this respiratory pattern.<br>Conclusion: Therefore, they are probably human. | Rule 1: If a patient presents with increased intracranial pressure (ICP), blurred vision, and nausea → infer high intracranial pathology.<br>Observation 1: The patient shows increased ICP, blurred vision, and nausea.<br>Rule 2: If CT scan shows no bleeding or swelling → rule out hemorrhagic causes.<br>Observation 2: CT scan reveals no evidence of bleeding or swelling.<br>Rule 3: If high ICP and hemorrhage is ruled out → suspect ischemic stroke.<br>Conclusion: Therefore, the patient most likely has an ischemic stroke. |
| Causal/ Counterfactual | Establishing a cause-and-effect relationship between events. | Ali, Muthu, Ah Hock, and Rentap exhibit this respiratory pattern. | A blood clot probably caused blockage in the brain leading to the stroke. |

### 4.3.1 Neurosymbolic Reasoning

Neuro-Symbolic AI (N-SAI) is an interdisciplinary field that aims to harmoniously integrate neural networks with symbolic reasoning techniques. Its overarching objective is to establish a synergistic connection between symbolic reasoning and statistical learning, harnessing the strengths of each approach. In the context of N-SAI, the symbolic system is used to represent predefined rulesets and knowledge bases, which then streamlines the process of making inferences and highlighting relationships between entities. Crucially, it is more transparent and more interpretable to humans. Meanwhile, the *neuro* component refers to artificial neural networks in the context of large-scale statistical learning. Artificial neural networks are adept at scale in classification, prediction and pattern recognition as well as processing unstructured data. Therefore, unifying and leveraging the strengths of both would hypothetically lead to the best of both worlds [54].

### 4.4 Trends in Existing Medical LLMs

While there is no shortage of LLMs applied to medical problems, there is a striking lack of methods which leverage *reasoning behaviour* in their operation [Table 2]. Comparing and contrasting this small subset of methods reveals some interesting trends.

First, inspecting their foundational or base models shows that unsurprisingly, most of these methods are built on generic LLMs, commonly GPT [55] or LLaMA [56] variants. This is likely due to their demonstrated effectiveness in day to day tasks, with more modern variants being shown to be surprisingly effective in clinical applications as-is [57]. However, it is notable that many approaches utilise multiple base models, with no single method relying on one model type. Relying on multiple models is unsurprising, as combining the strengths of multiple models is likely to boost overall effectiveness.

Second, most of their *reasoning behaviour* is derived from variants of chain-of-thought (CoT) [58] processes or reinforcement learning, likely because both techniques closely mimic cognitive processes fundamental to reasoning. CoT enables models to break down complex medical cases into a series of logical steps, mirroring the structured, stepwise reasoning that healthcare professionals apply. Additionally, few-shot learning complements CoT, allowing LLMs to "learn" from the input prompt, generalize from minimal clinical examples and adapt quickly to nuanced cases — a useful capability in medicine where data can be sparse or highly specialised. Meanwhile, reinforcement learning allows models to refine their reasoning capabilities through practice in a virtual simulation, improving accuracy through iterative feedback.

We note with interest that Supervised Fine-Tuning (SFT) and Reinforcement Learning from Human Feedback (RLHF) have been widely adopted to train medical LLMs for reasoning tasks, giving rise to several domain-specific models such as Huatuo GPT-o1 [59], MedR [6], MedVLM-R1 [5], and MedFound [60]. We refer to this class of systems as Large Reasoning Models (LRMs),a subclass of LLM, trained to perform extensive CoT reasoning. These LRMs employ a variety of RLHF strategies, including Proximal Policy Optimization (PPO) [37] for Huatuo GPT-o1 [59], Direct Preference Optimization (DPO) [38] for MedFound [60], and Group Relative Policy Optimization (GRPO) [39] for both MedR1 [6] and MedVLM-R1 [5]. Interestingly, findings from the MedR1 study revealed that models trained with GRPO not to output intermediate reasoning traces performed better in terms of final accuracy compared to those trained to explicitly generate intermediate steps [6]. This challenges the prevailing assumption that "more reasoning always leads to better outcomes," suggesting that, in some cases, compressed or implicit reasoning may yield higher task performance.

Third, the *reasoning behaviour* of most methods can be categorised as deductive reasoning, although there are a few cases of abductive and causal/counterfactual reasoning. Here, it is also worth noting that while LLMs excel in abductive reasoning tasks in multiple-choice scenarios, they are considerably less effective in generating hypotheses from scratch which may be of value in clinical use [61]. Since the overall goal of clinical diagnosis is to determine the disease affecting a patient from causative agents, the prevalence of deductive and to a lesser extent causal/counterfactual reasoning makes sense.

Finally, training datasets used vary widely in both scope and size, ranging across many different medical conditions, source material and between hundreds to thousands of samples. We observed no single standardised training dataset used by each approach, and as with architecture types many approaches used multiple training datasets. Most datasets were of the same modality (text data only), though some medical imaging datasets (MRI scans) were present. MIMIC-III was the most commonly used text dataset, with a combination of medical literature and other publicly available datasets [62]. Therefore, due to the differences in scope, strategy and data used by each approach, directly comparing *reasoning behaviour* across all models simultaneously is not presently feasible.

Aside from deductive reasoning, causal reasoning and neurosymbolic reasoning have also been demonstrably effective [Table 2]. However, example use cases are considerably less common. Current causal inference tests have a limited scope - such as determining the direction of causality between variable pairs, and their performance in more open-ended or nuanced causal inference as well as counterfactual reasoning remains unexplored. Meanwhile, neurosymbolic reasoning strategies exploit their inherently grounding properties to address the more fundamental issue of hallucinations in LLMs [72]. The diversity of strategies is striking - some methods exploit agent-based approaches to tailor argumentation schema and symbolic solvers for clinical reasoning [30], while others integrate dynamic medical ontologies in an attempt to more closely align *reasoning behaviour* with medical knowledge [73].

As each approach varies widely in scope and implementation, the advantages and disadvantages of each approach are broad. Generally, approaches using graph and decision tree-based strategies are easier to interpret due to their more deterministic nature, but may be less effective in ambiguous or complex cases (which are common in clinical practice). Meanwhile, methods which are more robust to noise or complex use cases are limited by a highly restricted scope, availability of training resources and a large computational footprint. Among these methods, DR HOUSE [2] is unique

Table 2: A table showing medical reasoning methods, their defining characteristics, and approach to reasoning.

| Method name | Base Architecture/Method | Reasoning Improvement Strategy | Type of Reasoning | Advantages | Disadvantages | Dataset | Github |
|---|---|---|---|---|---|---|---|
| Savage et al [4] | GPT-3.5; GPT-4.0 | Chain-of-Thought (Diagnostic Reasoning) | Deductive | Easy to implement | Scope is limited to GPT models, focusing exclusively on English medical questions | modified MedQA USMLE; NEJM (New England Journal of Medicine) case series | |
| Kwon et al [63] | GPT-4.0; OPT; LLaMA-2; 3D ResNet | Chain-of-Thought; Knowledge Distillation (via SFT) | Deductive | Lightweight and practical to use | Tight scope to limited disease conditions | Alzheimer's Disease Neuroimaging Initiative (ADNI); Australian Imaging Biomarkers and Lifestyle Study of Ageing (AIBL) | https://github.com/ktio89/ClinicalCoT |
| MEDDM Binbin et al [64] | GPT | Chain-of-Thought; Clinical Decision Trees | Deductive | Adaptable to different systems | Heavy data collection to generate clinical guidance trees | medical books, treatment guidelines, and other medical literature | |
| DRHOUSE [2] | GPT-3.5; GPT-4.0; LLaMA-3 70b; HuatuoGPT-II; MEDDM | Chain-of-Thought; Clinical Decision Trees | Deductive | Objective sensor measurement | Available datasets are currently limited | MedDG; KaMed; DialMed | |
| DR. KNOWS [65] | Vanilla T5; Flan T5; ClinicalT5; GPT | Chain-of-Thought; Extracted explainable diagnostic pathway | Deductive; Neurosymbolic | Hybrid method improves accuracy; Provides explainable diagnostic pathways | Particularly fragile to missing data | MIMIC-III; In-house EHR | |
| TEMED-LLM [66] | text-davinchi-003; GPT-3.5; logistic regression; decision tree; XGBoost | Few-shot learning, Tabular ML modelling; Neurosymbolic | Deductive | End-to-end interpretability, from data extraction to ML analysis | Requires human experts | EHR dataset (kaggle; see referenced publication for details) | |
| EHRAgent [8] | GPT-4 | Autonomous code generation and execution for multi-tabular reasoning in EHRs. | Deductive | Facilitates automated solutions in complex medical scenarios | Non-deterministic; Limited generalisability | MIMIC-III; eICU; TREQS | https://github.com/wshi83/EhrAgent; https://wshi83.github.io/EHR-Agent-page |
| AMIE [67] | PaLM 2 | Reinforcement Learning | Deductive | Effectively handles noisy and ambiguous real-world medical dialogues | Computationally expensive and resource-intensive; Simulated data may not fully capture real-world clinical nuances | MedQA; HealthSearchQA; LiveQA; Medication QA in MultiMedBench, MIMIC-III | |
| ArgMed-Agents [30] | GPT-3.5-turbo; GPT-4 | Chain-of-Thought; Symbolic Reasoning: Neurosymbolic | Deductive | Training-free enhancement; Explainability matches fully transparent, knowledge-based systems | Artificially restricted responses that do not match real-world cases | MedQA; PubMedQA | |
| Fansi Tchango et al [68] | BASD (Baseline ASD): Multi-Layer Perceptron (MLP) Diaformer | Reinforcement Learning | Deductive | Closely align with clinical reasoning protocols | Limited testing on real patient data | DDxPlus | https://github.com/mila-iqia/Casande-RL |
| MEDIQ [69] | LLaMA-3-Instruct (8B, 70B); GPT-3.5; GPT-4 | Chain-of-Thought; Information-Seeking Dialogues | Abductive | Robust to missing information | Available datasets are limited. Proprietary; Artificially restricted responses that do not match real-world cases | iMEDQA; iCRAFT-MD | https://github.com/stellalisy/mediQ |
| Naik et al [43] | GPT-4 | Causal Network Generation | Causal/ Counterfactual | Uses general LLMs | Lacks a specialised medical knowledge base | Providence St.Joseph Health (PSJH's) clinical data warehouse | |
| Gopalakrishnan et al [70] | BioBERT; DistilBERT; BERT; GPT-4; LLaMA | Causality Extraction | Causal/ Counterfactual | Easy to implement | Tight scope to limited disease conditions | American Diabetes Association (ADA); US Preventive Services Task Force (USPSTF); American College of Obstetrics Gynecology (ACOG); American Academy of Family Physician (AAFP); Endocrine Society | https://github.com/gseetha04/LLMs-Medicaldata |
| InferBERT [44] | ALBERT; Judea Pearl's Do-calculus. | Causal Inference using Do-calcus | Causal/ Counterfactual; Mathematical | Establishes causal inference | Tight scope to limited disease conditions; Highly restrictive input format | FAERS case reports from the PharmaPendium database | https://github.com/XingqiaoWang/DeepCausalPV-master |
| Emre Kıcıman [51] | text-davinci-003, GPT-3.5-turbo, and GPT-4 | Determine direction of causality between pairs of variables | Causal/ Counter-factual | Highly accurate for large models | Limited reproducibility due to dependency on tailored prompts | Tübingen cause-effect pairs dataset. | https://github.com/py-why/pywhy-llm |
| Huatuo GPT-o1 [59] | LLaMA-3.1-8B-Instruct and LLaMA-3.1-70B-Instruct | Supervised Fine-tuning and PPO | Deductive Reasoning | Instils multi-step reasoning in medical LLMs; Built-in interpretability as LLM can output reasoning traces along with answer | Limited evaluations as evaluations cover accuracy scores on medical MCQ benchmarks | Adapted from MedQA-USMLE and MedMcQA . | https://github.com/FreedomIntelligence/HuatuoGPT-o1 |
| Med-R1 [6] | Qwen2-VL-2B | Supervised Fine-tuning and GRPO | Deductive Reasoning | Joint image-text and multi-task reasoning; Built-in interpretability as LLM can output reasoning traces along with answer | Rethinking the "More Thinking is Better" Assumption | OmniMedVQA. | |
| MedVLM-R1 [5] | Qwen2-VL-2B | Supervised Fine-tuning and GRPO | Deductive Reasoning | Joint image-text reasoning; Built-in interpretability as LLM can output reasoning traces along with answer | Limited evaluations as evaluations cover accuracy scores on medical MCQ benchmarks | VQA-RAD, SLAKE, PathVQA, OmniMedVQA, and PMC-VQA | https://huggingface.co/JZPeterPan/MedVLM-R1 |
| MedFound [60] | 176-billion parameter LLM pretrained from scratch | Supervised Fine-tuning and DPO | Deductive Reasoning | - Self-bootstrapped Chain-of-Thought fine-tuning; Rigorous human evaluation of reasoning traces with rubric;Built-in interpretability as LLM can output reasoning traces along with answer | Proprietary EHR datasets aren't fully open, hindering exact reproduction | MedCorpus, MedDX-FT and MedDX-Bench | https://github.com/medfound/medfound |
| DeepSeekR1 [71] | DeepSeek-V3-Base | Supervised Fine-tuning and GRPO | Deductive Reasoning | Built-in interpretability as LLM can output reasoning traces along with answer | Pre-training and reasoning datasets aren't open-sourced | - | https://github.com/deepseek-ai/DeepSeek-R1 |

due to its EHR-free approach, only relying on objective sensor data to circumvent variance in clinical note interpretation. Unfortunately, code associated with many of these methods is not publicly available under an open-source licence, which limits our ability to inspect them in close detail. It is worth mentioning that medical LLMs are equally affected by some of the deeply rooted issues that plague general purpose LLMs as well, for instance memorisation [74] and hallucination [72].

# 5 Evaluating *reasoning behaviour* in Medical LLMs

To date, a standardised methodology for assessing the reasoning capabilities of large language models is absent. We review the current state-of-the-art in evaluation frameworks for analysing the *reasoning behaviour* of LLMs in medical tasks and we categorise evaluation methodologies into four distinct groups: (i) conclusion-based, (ii) rationale-based, (iii) interactive, (iv) and mechanistic evaluations [Figure **??**].

## 5.1 Conclusion-based Evaluation

In conclusion-based evaluation schemes, the focus is on the model's final answer rather than the reasoning process that led to it. Although this outcome-focused approach may overlook the model's underlying rationales, it can still offer valuable but limited insights into the model's reasoning patterns, especially if there is a clear cause and effect between different premises and conclusions in which the *reasoning behaviour* may be more-self evident.

| Evaluation Paradigm | Conceptual Focus | Typical Implementation & Metrics |
|---|---|---|
| **Conclusion-based** | Assesses correctness of the **final answer** only, without inspecting the reasoning path. | Automated scoring on Q-&-A benchmarks (e.g., MedQA, MedMCQA). Metrics: Accuracy, exact-match, F1. Fast, reproducible, but offers only high-level insight. |
| **Rationale-based** | Evaluates the **logic chain** or narrative explanation produced by the model. Focuses on **coherence, validity, and completeness** of reasoning traces. | Manual expert review or rubric-based grading of CoT. Automated graph checks (e.g., DAG similarity, causal-direction tests). Metrics: Bayesian Dirichlet score, Normalised Hamming Distance. |
| **Mechanistic** | Probes **low-level internal signals** to answer "*why did the model arrive here?*". Targets feature attribution and internal attention contributions. | Explainable-AI toolkits (Integrated Gradients, SHAP, attention rollout). Outputs saliency maps or keyword heat-maps for clinician inspection. |
| **Interactive** | Treats evaluation as a **dialogue or game**; dynamically stresses the model in real time. Explores the *response space* by challenging, re-prompting, or role-playing. | Game-theoretic tasks (e.g., debate, self-play). Rich insights but lower reproducibility; requires human-in-the-loop or scripted agents. |

Table 3: Comparison of reasoning evaluation paradigms in medical LLMs.

A wealth of benchmark data exists for the purpose of straightforward score-wise conclusion-based evaluation. A subset of simple benchmarks are available on the open medical LLM leaderboard, which cover question-and-answer tasks [10, 11, 12, 13]. Notably, some datasets consist of multiple-choice questions and some consist of short-answer questions. More sophisticated examples incorporating detailed data in multiple formats [75] or with more refined metrics also exist [76].

Inroads into more nuanced frameworks to gain deeper insight in *reasoning behaviour* for conclusion-based evaluation have been made. These take the form of frameworks evaluating paradigms at various levels, with high-level theoretical frameworks available [77]. A lower-level and detailed framework more suited for direct implementation is DR BENCH, which specifically assesses medical knowledge representation, clinical evidence integration and diagnosis accuracy [73]. In the process, an expanded suite of specific tasks is carried out to evaluate these three interdependent elements, and an accuracy score is reported for each sub-task to evaluate reasoning at a high level. However, we emphasise that conclusion-based evaluation is only capable of yielding high-level information due to its intrinsic nature.

### 5.2 Rationale-based Evaluation

In contrast to high-level conclusion-based evaluation schemes, rationale-based evaluation methods are process-driven instead of being outcome-driven. Their focus is on examining the reasoning process or *reasoning traces* generated by the model, typically assessing their logical validity and coherence. As rationale-based evaluation methods targeted at medical language models are relatively scarce and operate under distinct paradigms, we will discuss them individually on a case-by-case basis.

The most straightforward but resource-heavy approach was to manually evaluate answers using the skills of domain experts. These domain experts were blinded to the questions and identified logical fallacies as well as inaccuracies directly in provided rationale [4]. More structured variants of this method have emerged, where expert clinicians employ clinically validated rubrics to score LLM-generated reasoning traces. Notable examples include the CLEVER (CLinical EValuation for Effective Reasoning in Diagnosis) rubric [60], and the Revised-IDEA (R-IDEA) rubric [78, 79].

Furthermore, clinicians have conducted deeper analyses of models such as DeepSeekR1 using USMLE-style multiple-choice questions, uncovering several notable failure modes. These included: anchoring on initial symptoms while disregarding contradictory findings, omission of standard-of-care treatments, misuse of laboratory values or drug mechanisms, and confusion between clinically similar entities (e.g., troponin vs. CK-MB) [80]. Of particular interest was a failure mode termed Chain-of-Thought (CoT) mismatch [80], where the reasoning path supported one answer, but the model selected a different final option [80]—highlighting that reasoning traces, while useful, are not a perfect proxy for model interpretability or correctness. We note the value and effectiveness of clinician-led evaluations of LLM reasoning traces, as they offer nuanced insights into clinical reasoning quality. However, these methods are inherently time-consuming and reliant on the availability of expert clinicians.

To reduce the reliance on time-intensive clinician reviews, recent work has turned to automated reasoning-trace evaluation [81, 82]. When a high-quality “gold-standard” CoT exists, generated traces can be compared step-by-step using text-similarity metrics such as BLEU, METEOR, or BERTScore [81]. Additionally, a secondary “judge” LLM can be prompted to assess another model’s chain of thought—either by comparing it against a provided standard [82, 81, 83] or in the case where no references exist the "judge" LLM can be prompted to evaluate for logical coherence and clinical relevance based on its own internal reasoning [81]. Another LLM evaluation workflow integrated online search to fact check reasoning traces [83].

Conversely, an automated approach applied Directed Acyclic Graphs (DAG) to represent underlying relationships in complex medical datasets, including cancer [84]. In implementation, a DAG was constructed by predicting which factors might influence others, and accuracy was scored with a Bayesian Dirichlet metric measuring the similarity of the resultant graph with the ground truth of real patient data. In addition, a separate method also applied DAG, but exploited it to infer the direction of causality between variable pairs [42]. Accuracy was then measured using Normalised Hamming Distance (NHD) as a similarity metric between the resultant and ground truth patient outcome or diagnostic graph.

### 5.3 Mechanistic Evaluation

Similarly, mechanistic evaluation of *reasoning behaviour* is process driven with the aim of examining low-level reasoning traces. In contrast to rationale-based evaluation, mechanistic evaluation delves deeper into the underlying processes that drive a model’s response, aiming to uncover the fundamental questions of "how" and "why" associated with an outcome.

In practice, feature attribution methods can be exploited to study *reasoning behaviour* by highlighting keywords which are conceptually identical to features of interest in medical LLMs. These explainable AI (XAI) methods compute an attribution score for each input feature to represent its contribution to a model’s prediction, which can be calculated and represented with a variety of metrics [19, 85]. For example, a hypothetical medical scenario may show that the key words "blocked nose" are strongly weighted in a positive influenza prediction. In this context, the key word is equivalent to the reasoning trace, and is shown to impact the model’s *reasoning behaviour*. A conceptually similar strategy has been applied to explain predicted diseases from patient-doctor dialogues [86].

### 5.4 Interactive Evaluation

Finally, a more open-ended approach to reviewing *reasoning behaviour* is interactive evaluation. Unique to other strategies, it is reactive and engages with the LLM directly during evaluation, adjusting questions to fit the model’s response. This deeper exploration of the "response space" tests and further exposes the model’s reasoning capabilities as well as limitations [87]. Variants of interactive evaluation may challenge the model’s conclusions directly [88] or use game-theoretical scenarios to probe reasoning depth [89].

One notable implementation of this paradigm is Sequential Diagnosis Benchmark (SDBench) [3]. Most existing medical LLM benchmarks present models with all clinical facts at once and assess multiple-choice accuracy—conditions far removed from real-world diagnostic workflows. To address this, SDBench [3] introduces a simulation-based framework that converts 304 NEJM Clinicopathological Conference (CPC) cases into interactive diagnostic scenarios. Here, the LLM or physician must sequentially ask patient history questions, order diagnostic tests (each with an associated monetary cost), and commit to a final diagnosis [3]—closely mimicking real clinical cases under uncertainty and cost constraints. This setup not only enables evaluation of diagnostic accuracy but also the diagnostic yield of tests as measured by total cost per case. By transforming passive vignette-based evaluation into a structured, decision-based simulation, SDBench enhances realism while maintaining standardisation and reproducibility within interactive evaluation. However, it is not without limitations: the CPC dataset is biased toward complex cases where there are no healthy patients in the medical cases, test costs are drawn only from U.S. price list, and benchmarking physicians were restricted from using web search to prevent them from finding the original CPC cases [3].

Unfortunately, a critical flaw of this evaluation method is its lack of reproducibility and standardisation due to its reactive nature. Currently, one exception exists, circumventing irreproducibility by side-stepping the requirement for a prompt [90]. Nevertheless, we note the strong advantages of interactive evaluation, and note that it remains relatively unexplored in the current medical LLM literature. Refinement of the core method and further investigation of strategies (such as the aforementioned prompt skipping [90]) to counteract its limitations have the potential to raise its reproducibility to reasonable levels.

### 5.5 Summary of Evaluation Strategies

To our surprise, we found few existing studies of *reasoning behaviour* evaluation in a medical LLM context. Nevertheless, we note several broad insights from the few existing studies matching our scope: (a) graph-theoretic approaches are intuitively applicable to evaluating causal or counterfactual *reasoning behaviour* due to their representation of cause-and-effect, (b) feature attribution methods provide a low-level glimpse into medical reasoning, (c) interactive evaluations like SDBench [3] transforms static multiple-choice medical benchmarks into interactive simulations that more realistically reflect clinical reasoning under uncertainty and resource constraints, (d) while manual clinician evaluations of reasoning traces offer nuanced insights that traditional metrics miss but they are difficult to scale due to time and expertise requirements; and (e) *reasoning behaviour* evaluation methods are complementary, with the potential of being applied simultaneously to obtain a better understanding in cases where the model configuration allows.

## 6 Towards transparency in medical LLMs

Given our findings, we pose the central question: **How can we develop methods that expose the underlying *reasoning behaviour* in medical LLMs?**

To answer this, we introduce two conceptual and theoretical frameworks intended to expose reasoning behaviour in medical LLMs with orthogonal design principles. These frameworks are designed to meet two important criteria: (a) low-level *reasoning behaviour* must be visible and the framework should be (b) task-agnostic. Each framework would consist of three broad stages: (a) data preprocessing, (b) model training and (c) interpretability via extraction of *reasoning behaviour*. These designs—one straightforward and one relatively more complex—are detailed in the following subsection, **Theoretical Frameworks for Transparent Reasoning Behaviour**, and visualized in Figure 6.

Additionally, we observe with interest the growing prominence and impressive capabilities of LRMs such as Anthropic's Claude 3.7 Sonnet, Google's Gemini 2.5, and OpenAI's o1. Similarly, these models demonstrate key attributes aligned with our framework criteria: (a) visibility into low-level reasoning behavior that supports interpretability, and (b) task-agnostic flexibility across domains. Although the precise details of their training pipelines remain undisclosed, these models likely rely on a combination of extensive supervised fine-tuning on curated reasoning datasets, large-scale reinforcement learning for reasoning alignment, and inference-time scaling techniques. These approaches have enabled such models to excel on existing benchmarks from math and logic benchmarks to coding benchmarks [71, 91, 92]. Open-source alternatives such as DeepSeek R1 [71] and Alibaba's Qwen 2.0 [91], Qwen 2.5 [92], as well as domain-specialized models like MedVLM-R1 [5], MedR1 [6], and MedFound [60], have adopted broadly similar strategies tailored to medical contexts. In the subsequent subsection, **Open Challenges in Large Reasoning Models**, we outline several open challenges that constrain further progress in this space.

### 6.1 Theoretical Frameworks with more transparent *reasoning behaviour*

We begin the simplistic framework by restricting input data scope to standardised data formats. To this end, TEMED-LLM [66] can be used to parse textual data into tables in the preprocessing stage with a predetermined format. Structured input has multiple advantages, being consistent and more easily ingested into software. In addition, a side effect is further simplification of data. An advantage of this which may benefit machine learning algorithms is a noise reduction while increasing variance in the data. However, we note that a degree of low-level feature loss is possible. Next, we consider that while deep learning is a powerful tool, more conventional machine learning approaches are often sufficient in many cases. We exploit the tabular nature of the data and leverage tree-based methods, which include examples such as XGBoost [93] or Random Forest [94]. While straightforward, these are effective and particularly suited to p » n problems common across the biological sciences, where there are far more features per sample than there are samples, i.e. the "curse of dimensionality" [95]. In addition, tree-like approaches have additionally benefited from properties that make them more interpretable. Exploiting this property allows us to generate decision sets which are interpretable during model training [96], hence exposing *reasoning behaviour*.

Our second proposed framework requires specific context. We refer to the categorisation of reasoning into two systems of thinking [97] [Figure 7]. "System 1" thinking refers to more instinctive decision-making based on learned patterns and experiences. Although conventional LLMs often fall into this category, they perform effectively given their vast input corpora, similar to a human memorising vast quantities of data. Conversely, "System 2" thinking refers to more thorough and conscious thinking used in problem-solving, and requires relatively more effort to achieve, both in humans and machine learning. Neither system operates fully independently, where the rapid assessments in "System 1" thinking form the foundation for more methodical "System 2" thinking. We note that we neither support nor disregard this overall viewpoint, but find value in using this angle to frame our proposed strategy, which addresses "System 1" limitations inherent to conventional LLMs.

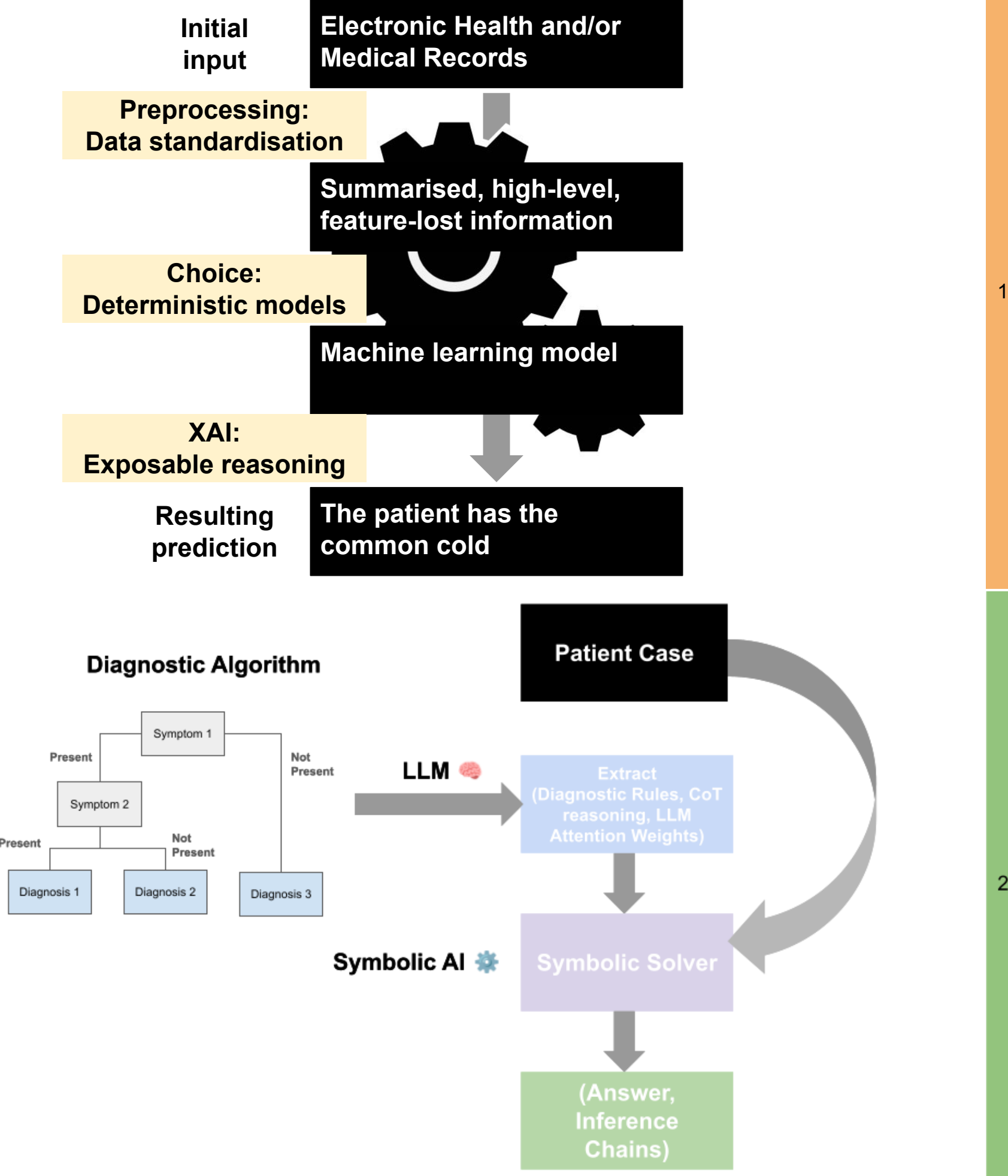

Figure 6: Two frameworks with a focus on exposing *reasoning behaviour*. Note that the two frameworks are independent but shown together to facilitate easier comparison. **Top: Input data is standardised and fed to tree-based models.** The deterministic nature of trees are exploited for achieving transparency for *reasoning behaviour*. **Bottom: An integrative framework of combining the complementary strengths of LLM and Symbolic Reasoning.** The medical LLM extracts diagnostic rules from clinical algorithms, along with its Chain-of-Thought (CoT) reasoning and attention weights. These diagnostic rules, together with patient case inputs, are provided to the symbolic solver, which determines the final diagnosis and generates inference chains as its reasoning trace.

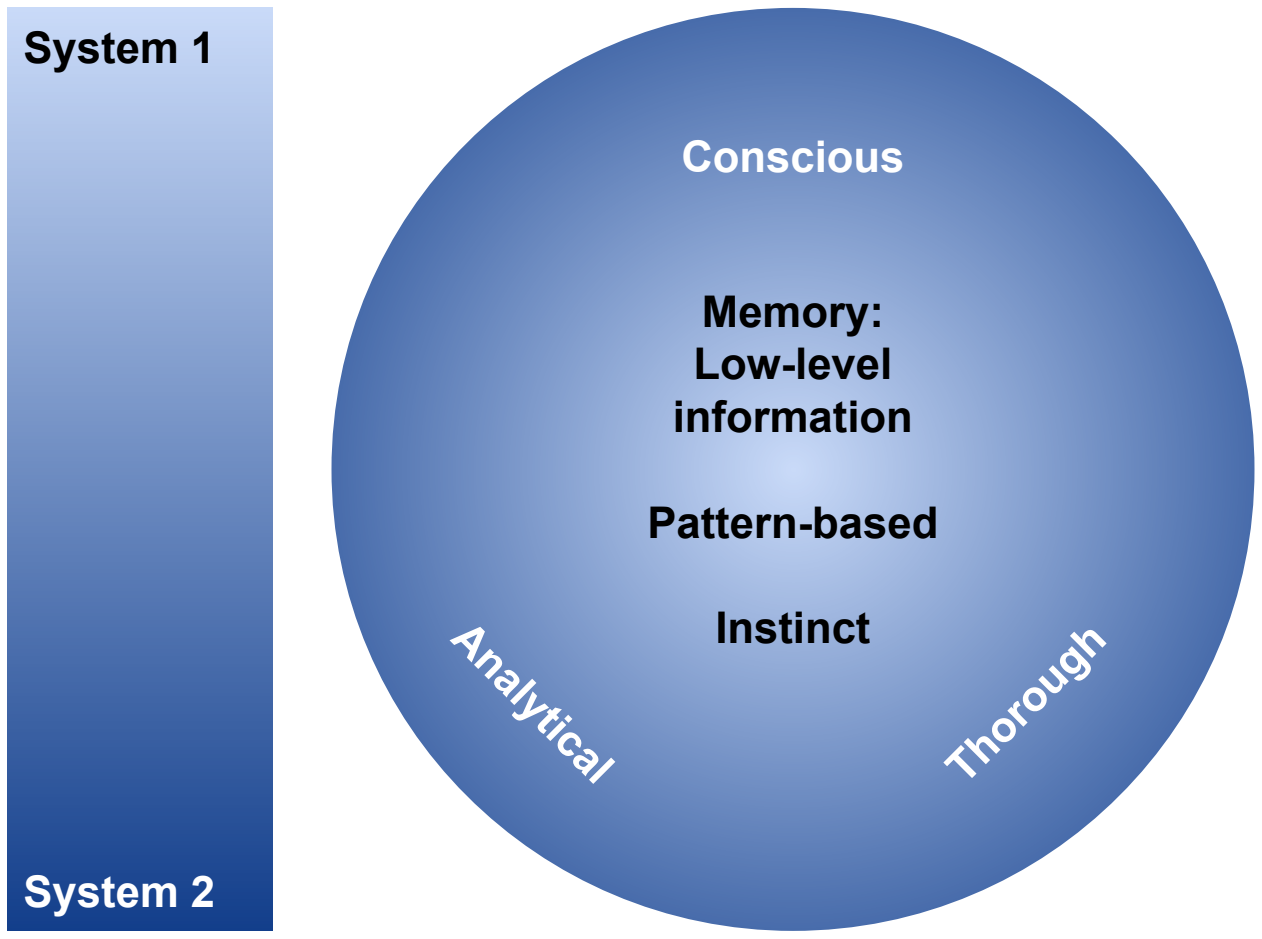

Figure 7: An illustration of the spectrum of "System 1" fundamental thought processes to "System 2" analytical thought processes.

Next, we observe with interest that Symbolic AI (SAI) incorporating symbolic reasoning excels in "System 2" thinking. In contrast, SAI has limited performance in "System 1" thinking, lacking the capacity for rapid, intuitive pattern recognition and memorisation, hindering its performance in tasks where large volumes of unstructured data are processed. In order to utilise SAI more effectively, a more defined knowledge representation representing the search space with an optimal solution is often required.

Therefore, our final proposed framework synthesises LLMs and SAI, leveraging their complementary traits to supplement each other. The methodology consists of three core stages: (a) data preprocessing, (b) parallel model integration and (c) reasoning extraction. One would first aim to generate a structured dataset which can be leveraged by LLM and SAI in data preprocessing, considering that many medical datasets contain highly unstructured data. Next, both LLM and SAI would be implemented. The purpose of the LLM would be to form a knowledge base capable of generating initial hypotheses, suggesting pertinent medical literature and proposing logical diagnostic rules inferred from clinical practice guidelines. Meanwhile, one would develop a formal SAI reasoning system encoding medical knowledge, clinical guidelines and diagnostic rules. By applying logical reasoning to patient data, conclusions would be consistent with established medical protocols. Finally, to trace *reasoning behaviour* it is possible to extract both attention weights and CoT-generated explanations from LLMs, as well as output logical inference chains from SAI. Subsequently combining the reasoning traces provides a more comprehensive glimpse into how the final conclusions were reached for medical professionals. Similar to its use in our second proposal, the PRM can be trained on reasoning traces— CoT explanations from the LLM, and inference chains from SAI—to evaluate reasoning behavior. More specifically, the complementary nature of (a) assessing the logical coherence of these traces and (b) identifying potential logical fallacies lend additional credibility to the combined reasoning outputs.

## 6.2 Open Challenges in Large Reasoning Models

### 6.2.1 Reasoning Data and Evaluation

**Reasoning-knowledge entanglement**
Current benchmarks conflate reasoning ability with domain knowledge, reflecting a tension between intuitive, memory retrieval-based (System 1) and analytic (System 2) reasoning processes. It remains challenging to determine whether the performance of a model is the result of genuine logical reasoning or simply retrieval of memorized facts with some generalization from what was memorized. Evaluating genuine reasoning performance requires benchmarks intentionally designed to minimize prior core knowledge dependency, to isolate reasoning capabilities from knowledge recall. The **ARC-AGI** ("Abstraction and Reasoning Corpus" for Artificial General Intelligence) benchmark exemplifies this approach, assessing how efficiently an AI can abstract and generalize from limited input on visual tasks with minimal knowledge priors, relying on *reasoning* rather than memorised content [98, 99]. To the best of our knowledge, no equivalent benchmark currently exists for the medical domain.

**Limited Availability of High-Quality Reasoning Traces**
Chain-of-Thought (CoT) data are critical for training reasoning models, yet richly annotated, diverse medical CoT corpora remain relatively scarce and costly to produce with human annotators (i.e. clinicians). One promising remedy is to harness advanced LLMs themselves to generate synthetic CoT examples: their long context windows and fine-grained instruction-following abilities enable them to produce detailed, multi-step rationales on demand.

**Evaluation Metrics Beyond Accuracy**
Evaluating reasoning behavior in medical LRMs requires moving beyond simple accuracy metrics, which often fail to capture the quality and structure of clinical reasoning processes. Instead, the focus should shift toward metrics that assess reasoning trace fidelity—such as logical soundness, coherence, and completeness. This approach is supported by recent work on CoT monitorability [100], which argues that CoTs offer a fragile but valuable window into model reasoning. Integrating CoT monitoring into model oversight enables early detection of incorrect or unsafe reasoning patterns, an important consideration in high-stakes domains like medicine.

Rubrics like CLEVER [60] and R-IDEA [78, 79], which have been developed to evaluate clinical reasoning in humans, offer a promising template for benchmarking LRMs. By scoring LRM-generated chain-of-thoughts against these rubrics and comparing them directly with evaluations of human clinicians on the same tasks, we can derive more nuanced and clinically relevant assessments of LRM reasoning performance.

While the use of 'judge' LRMs offers a scalable path for automated reasoning trace evaluation—discussed in detail in the rationale-based evaluation section—it is important to acknowledge their limitations, particularly concerning bias [101]. These models inherit patterns from their training data, which may embed societal stereotypes associated with race, gender, religion, culture, and ideology [101]. Evaluating the robustness and fairness of such systems in medical contexts remains an essential direction for future work.

### 6.2.2 Training and Inference Optimisations

**Scalability of RL-based approaches**
Reinforcement learning (RL) methods such as PPO [37], DPO [38], and GRPO [39]have shown strong potential in improving LRM reasoning capabilities. However, these approaches are computationally intensive and require significant resources, which can hinder widespread adoption. The key challenge is to develop RL-based reasoning techniques that are less computationally intensive without compromising performance.

**Medicine/Healthcare specific reward models or reward functions**
The generalisability of reward models and functions developed for broad-domain LRMs to medical reasoning tasks remains uncertain, given the domain-specific nuances of clinical decision-making. One promising direction involves incorporating high-quality evaluations of reasoning traces directly into the reward modeling process, thereby enhancing reasoning alignment. This approach aligns with the principles of process supervision [102], which has been shown to outperform outcome supervision—particularly in tackling complex problems in mathematics [102, 103]. While process supervision traditionally depends on labor-intensive human annotation of step-by-step reasoning, this limitation can be addressed through Process Reward Models (PRM) [102]: customised reward systems trained on annotated Chain-of-Thought (CoT) traces to automate reasoning evaluation during RLHF. These models assess logical coherence and detect fallacies, preserving transparency without requiring constant human input. An intriguing extension is to integrate LRM-based reasoning trace evaluators as those covered in **Rationale-based Evaluations** directly into the RLHF pipeline, enabling scalable process supervision as explored in the rationale-based evaluation paradigm.

**Scaling laws for LRM inference**
Studies have demonstrated that increasing compute during inference-time—rather than solely during training—can significantly enhance the reasoning performance of LRMs [31, 32]. Techniques like Chain-of-Thought (CoT) and Tree-of-Thoughts (ToT) [26] allow models to generate multiple intermediate reasoning paths, which can then be evaluated or expanded upon. This scaling behavior is further supported by methods such as self-reflection [32], Monte Carlo Tree Search (MCTS) [32], and Process Reward Models (PRMs) [31], which help verify and select coherent reasoning sequences during inference. Additionally, integrating fact-checking mechanisms via web-enabled LLMs adds a layer of external validation, enabling more accurate and trustworthy output. Together, these approaches illustrate how smarter use of inference-time compute can unlock deeper reasoning capabilities without retraining.

### 6.2.3 Alternative Reasoning Architectures

LLMs and subsequently LRMs, rely on left-to-right, next-token prediction, limiting their capacity to plan, revise, or backtrack during inference. This property occurs as a result of the autoregressive nature of generative LLMs, which produce an output token by selecting a token with the highest probability score based on previously generated tokens. Should a model choose one "nonsense" token in context, subsequent tokens are similarly affected. A compounding effect can occur, quickly worsening the error unless the model has the ability to backtrack. To achieve more structured and accurate reasoning, alternative architectures that emphasize planning-based non-autoregressive strategies, such as JEPA [104], warrant further exploration. Another promising alternative is latent reasoning[105, 106], where models perform internal, iterative computations on hidden states before outputting tokens, enabling internal planning, error correction, and revision prior to generating any output.

#### 6.2.4 Multimodal Clinical Reasoning Integration

Clinical data is inherently multimodal, involving the integration of information across text, medical imaging, electronic health records (EHRs), sensor data (e.g., from smartwatches), and physiological time-series such as ECG or EEG. The ability of LRMs to unify understanding and reasoning across these diverse medical modalities within a single cohesive system indicates an exciting direction for future research. To advance LRM's multimodal reasoning capabilities, we require complex reasoning data synthesis pipelines that go beyond the current paradigm, which largely focus on single modalities. The path forward involves aligning multiple modalities simultaneously and building rich, interleaved reasoning chains that reflect how clinicians synthesise diverse data sources to arrive at diagnostic and treatment decisions.

## 7 Discussion

### 7.1 A striking lack of evaluation methods for Reasoning Behaviour

As part of our study, we intended to investigate the current SOTA of medical-LLM performance in the context of *reasoning behaviour*. However, we found that only a few existing evaluation frameworks adequately capture the nuances involved in assessing *reasoning behaviour* in medical LLMs. This is perhaps unsurprising, given the inherent complexity of such evaluations—particularly those requiring human clinicians. To manually assess reasoning traces of LLMs is time-consuming and difficult to scale.

While conventional conclusion-based evaluations are not without value, they offer only a limited view of *reasoning behaviour*, as they focus solely on final answers without analysing the reasoning process. Evaluating reasoning traces in large language models (LLMs) is crucial, especially in medical contexts, as it is possible for models to reach the correct conclusion through erroneous reasoning. Therefore, an AI that not only gives a diagnosis but also provides a corresponding rationale behind it can significantly aid adoption among clinicians and other healthcare professionals by virtue of its transparency which fosters trust in its recommendations. Broadly, we consider that rationale-based, interactive and mechanistic evaluation are more naturally predisposed to decrypting the *reasoning behaviour* of medical LLMs. Interactive evaluations such as the Sequential Diagnosis Benchmark (SDBench) offers a promising alternative to static MCQ benchmarks, by simulating interactive diagnostic scenarios that better mirrors real clinical workflows.

### 7.2 Memorisation vs Planning: The "Stochastic Parrot" Problem

Another potential issue is the lack of memorisation tests and benchmarks in LLM. This is pertinent as (like humans) medical LLMs have the ability to memorise the dataset they are given but on a much grander scale, giving the illusion of reasoning although in reality regurgitating related or unrelated information from a vast knowledge base (like humans) [74]. Hence, in many cases it was not possible to answer the question: **"to what extent is this model a stochastic parrot and to what extent is this model performing logical reasoning?"** To answer this question, a structured approach would involve testing its ability to work from foundational first-principles, or "base facts," embedded within its training corpus. These base facts encompass simple yet essential principles across areas such as medicine, physiology, and pharmacology, often representing core medical knowledge. For example, a base fact in medicine could be: "The heart pumps blood throughout the body." Reasoning tests can then be designed to see if the model can apply such base facts to answer more complex questions.

In practice, however, access to high-volume training corpora for closed-source enterprise models like GPT-4 [55], Gemini [107], or Anthropic [108] is restricted. This limitation calls for designing medical tests that embed low-level fundamental knowledge, and relying primarily on the model's ability to reason from these base facts. Nevertheless, we do not intend to diminish the usefulness of "System 1" rote-memorisation type LLMs (which are a prerequisite for "System 2" advanced-reasoning type systems), but instead wish to highlight the lack of insight into a model's *reasoning behaviour* without this layer of validation.

### 7.3 Hallucination and Security Implications of Opaque Reasoning

A natural side effect of obtaining transparency into medical *reasoning behaviour* is its neutralising effect on hallucination events common across LLMs in all domains [72]. Such events occur due to the autoregressive nature of generative LLMs, which produce an output by selecting a token with a highest probability score. Should a model choose one "nonsense" token in context, subsequent tokens are similarly affected. A compounding effect can occur, quickly worsening the error unless the model has the ability to backtrack. However, exploring *reasoning behaviour* allows greater insight into hallucination events by exposing the involved logic chain, complementing the current state of the art of retrieval-augmented generation (RAG) [109].

More ominously, it is an unfortunate reality that cybercrime is increasingly common, and it is not impossible that healthcare infrastructure, including associated medical LLMs may be targeted. While this commonly takes the form of ransomware, it takes disturbingly little effort to "poison" medical LLMs with misinformation [110], with myriad implications for those used for clinical diagnosis or hospital operations. This aspect of medical LLMs is a relatively unexplored field, with most studies focusing on generic use cases [111], though a more comprehensive framework incorporating the paradigms discussed in our review as well as this security aspect exists for medical vision language models [112]. As with exposing hallucination events, greater transparency into medical models will assist in identifying such events should the situation arise.

### 7.4 Reasoning Behaviour as a Form of Explainable AI

Enhanced medical reasoning transparency may contribute to solving on-going problems such as addressing differential diagnosis or providing clinical management plans. Differential diagnosis is an on-going problem in medical AI development, where similar conditions may confound a prediction with potentially severe consequences. By viewing reasoning traces, both method developers and clinicians will be able to better discern why an accurate or inaccurate choice is made, and adjust the model accordingly as well as gaining potentially unknown clinical insights.

**Finally, we highlight the point that understanding *reasoning behaviour* of medical LLMs is functionally equivalent to achieving explainable AI (XAI),** and is not mutually exclusive with other XAI techniques or evaluation paradigms discussed. Given the understandably high level of scrutiny placed on medical methods, achieving this deeper level of understanding is necessary to demonstrate the effectiveness of medical LLMs. At the same time, we may be able to answer an interesting question: "is improved reasoning correlated to improved performance?" Ultimately, understanding the *reasoning behaviour* of medical LLMs and LLMs in general has applications across all domains - especially since LLMs are also effective in computer vision tasks.

## 8 Conclusion

In summary, it is intuitive that a greater understanding of the *reasoning behaviour* of medical LLMs empowers clinicians, improves patient trust in them and allows machine learning engineers to troubleshoot underperforming models. However, the lack of studies focusing on understanding *reasoning behaviour* is striking, where the majority of studies are focused on achieving high-level performance metrics with a conspicuous lack of focus on XAI. Most *reasoning behaviour* evaluation strategies are in their infancy, though there is notable potential for growth and further studies. Our theoretical proposed frameworks, while limited, can contribute to XAI in clinical LLMs, with the ultimate goal of improving transparency in medical AI and subsequently patient outcomes.

## 9 Abbreviations

CoT: Chain of Thought CLEVER: CLinical EValuation for Effective Reasoning in Diagnosis DAG: Directed Acyclic Graphs LLM: Large Language Models MCTS: Monte Carlo Tree Search NHD: Normalised Hamming Distance N-SAI: Neuro-Symbolic AI PRM: Process Reward Models R-IDEA: Revised-IDEA ToT: Tree of Thought

## 10 Declarations

### 10.1 Ethical Approval

Not applicable.

### 10.2 Consent for publication

Not applicable.

### 10.3 Competing Interests

The authors declare that they have no competing interests.

### 10.4 Funding

Not applicable.

### 10.5 Author's Contributions

Conceptualization, both authors; formal analysis, both authors; funding acquisition, both authors; investigation, both authors; resources, both authors; supervision, T.C.; validation, both authors; visualization, both authors; writing—original draft, both authors; writing—review and editing, both authors.

## 11 Acknowledgements

We thank Eric Lee Kuan Hui for facilitating the initial meeting between the authors and subsequent collaboration. We thank Jim Liew Jun Fei, Pui Wee Yang and Winston Lee Meng Yih for critically reviewing the manuscript.